\documentclass[sigconf]{acmart}

\usepackage{xcolor}

\def\BibTeX{{\rm B\kern-.05em{\sc i\kern-.025em b}\kern-.08emT\kern-.1667em\lower.7ex\hbox{E}\kern-.125emX}}
    
\copyrightyear{2019}
\acmYear{2019}
\setcopyright{rightsretained}
\acmConference{Preprint}{July 2019}{}

\begin{document}

%
\title[Addressing Delayed Feedback for Continuous Training in CTR prediction]{Addressing Delayed Feedback for Continuous Training with Neural Networks in CTR prediction}
%

%

\author[1]{Sofia Ira Ktena}
\affiliation{%
  \institution{Twitter}
  \city{London}
  \country{UK}
  }
\email{siraktena@twitter.com}

\author{Alykhan Tejani}
\affiliation{%
  \institution{Twitter}
  \city{London}
  \country{UK}
  }
\email{atejani@twitter.com}

\author{Lucas Theis}
\affiliation{%
  \institution{Twitter}
  \city{London}
  \country{UK}
  }
\email{ltheis@twitter.com}

\author{Pranay Kumar Myana}
\affiliation{%
  \institution{Twitter}
  \city{London}
  \country{UK}
  }
\email{pmyana@twitter.com}

\author{Deepak Dilipkumar}
\affiliation{
  \institution{Twitter}
  \city{San Francisco}
  \country{USA}
}
\email{ddilipkumar@twitter.com}

\author{Ferenc Husz\'{a}r}
\affiliation{%
  \institution{Twitter}
  \city{London}
  \country{UK}
  }
\email{fhuszar@twitter.com}

\author{Steven Yoo}
\affiliation{
  \institution{Twitter}
  \city{San Francisco}
  \country{USA}
}
\email{syoo@twitter.com}

\author{Wenzhe Shi}
\affiliation{%
  \institution{Twitter}
  \city{London}
  \country{UK}
  }
\email{wshi@twitter.com}

%
\renewcommand{\shortauthors}{Ktena et al.}

%
\begin{abstract}
One of the challenges in display advertising is that the distribution of features and click through rate (CTR) can exhibit large shifts over time due to seasonality, changes to ad campaigns and other factors. The predominant strategy to keep up with these shifts is to train predictive models continuously, on fresh data, in order to prevent them from becoming stale. However, in many ad systems positive labels are only observed after a possibly long and random delay. These delayed labels pose a challenge to data freshness in continuous training: fresh data may not have complete label information at the time they are ingested by the training algorithm. Naive strategies which consider any data point a negative example until a positive label becomes available tend to underestimate CTR, resulting in inferior user experience and suboptimal performance for advertisers. The focus of this paper is to identify the best combination of loss functions and models that enable large-scale learning from a continuous stream of data in the presence of delayed labels. In this work, we compare 5 different loss functions, 3 of them applied to this problem for the first time. We benchmark their performance in offline settings on both public and proprietary datasets in conjunction with shallow and deep model architectures. We also discuss the engineering cost associated with implementing each loss function in a production environment. Finally, we carried out online experiments with the top performing methods, in order to validate their performance in a continuous training scheme. While training on 668 million in-house data points offline, our proposed methods outperform previous state-of-the-art by 3\% relative cross entropy (RCE). During online experiments, we observed 55\% gain in revenue per thousand requests (RPMq) against naive log loss.
\end{abstract}

%
%


%

%

%
\maketitle

\section{Introduction}

\begin{figure}[t]
    \centering
    \includegraphics[width=.9\linewidth]{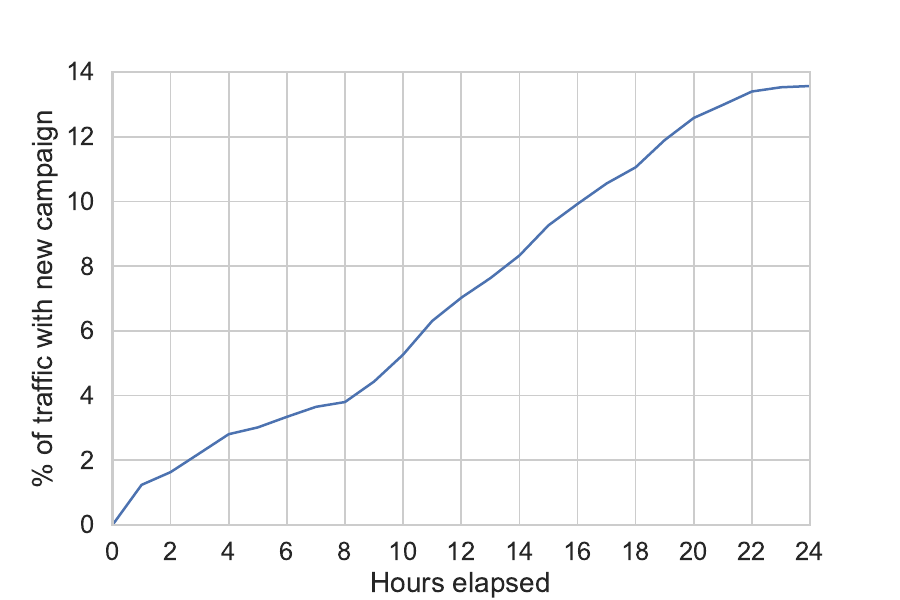}
    \caption{Percentage of traffic with new campaign IDs for each hour following a reference day.}
    \label{fig:campaign_ids}
\end{figure}

\begin{figure}[t]
    \centering
    \includegraphics[width=0.9\linewidth]{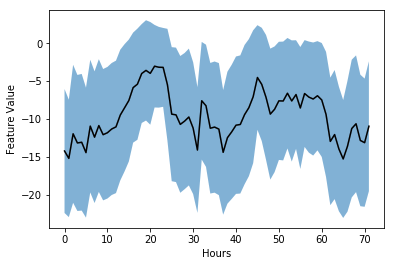}
    \caption{Distribution of a continuous feature's value across 3 days.}
    \label{fig:continuous_feature}
\end{figure}

\begin{figure}[t]
    \centering
    \includegraphics[width=\linewidth]{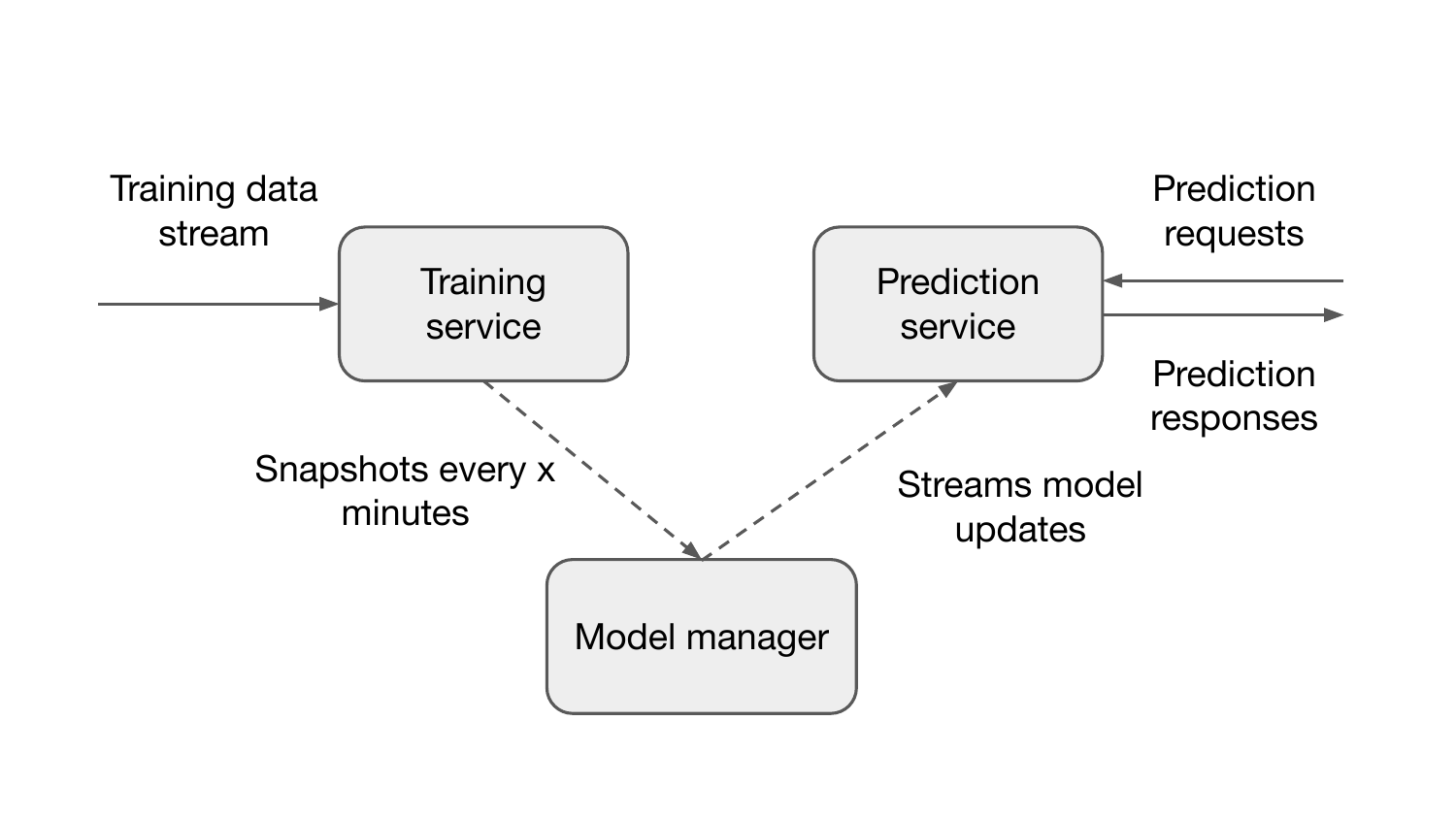}
    \caption{Continuous training framework.}
    \label{fig:continuous_training}
\end{figure}

Many advertisers would choose not to pay for an impression unless a user takes a predefined action, such as clicking on their ad or visiting their website. This issue is well-known in the field of advertising and models such as cost-per-click (CPC) and cost-per-conversion (CPA) were introduced to allow advertisers to only pay for predefined engagements. These performance-dependent payment models require the estimation of the probability that an impression will lead to a specific engagement. 
One of the challenges in display advertising is that feature and click through rate (CTR) distributions can experience big shifts due to special events, new campaigns and other factors. Figs.~\ref{fig:campaign_ids},~\ref{fig:continuous_feature} illustrate these changes through time. To address this problem, Twitter's ads prediction models are constantly trained online on fresh data, i.e. they take a continuous stream of data and update the model when new data arrives (see Fig.~\ref{fig:continuous_training}). 

A major challenge encountered in the above scenario is that of delayed feedback of user actions. Some engagements, like a click on the ad or an MRC view\footnote{\url{http://mediaratingcouncil.org}} of the ad video, may occur with a time delay of 1 minute, 1 hour or even 1 day after the ad is displayed. The challenge that arises then is whether to wait for a fixed time window before assigning a label to ad impressions, and subsequently train on the data, or rather decide on the label based on certain heuristics. The main downsides of the former approach are that the model can become stale while waiting and the additional infrastructure cost of maintaining a data cache. In the latter case, training entails falsely labeling examples as negatives, resulting in more negative examples than in the actual data distribution. It is also unclear what the ideal window length would be, in order to find a trade-off between the delay in model training and the fake negative (FN) rate (i.e. incorrectly labeling examples as negative due to the window being too short).

Internally, empirical results show that even a 5-minute delay to model updates can prove to be very damaging in terms of performance. 
In the naive continuous learning approach, small batches of training samples are immediately ingested with negative labels, hence the model is always up-to-date, even in the case of a severe feature distribution shift across time. Because there is no need to store additional snapshots of examples that have not been clicked yet, this approach technically allows us to wait for an infinite amount of time until a positive label is observed. Then, it can be immediately introduced to the model, as shown in Figure~\ref{fig:continuous_training}. The main drawback of this approach is that all samples are initially labeled as negatives, even though the user might eventually engage with them. Alternatively, if the data samples remain unlabeled until an engagement takes place and the probability of engagement is considered to be dependent on the time elapsed since impression as in~\cite{chapelle2014modeling}, regular snapshots of the data need to be collected and stored to capture the engagement delay distribution. Although the assumption of time-dependence is valid, this would lead to a large increase in infrastructure cost given that data points need to be stored multiple times (once for each snapshot) to train the model in batch mode, causing the training data to grow overtime without aggressive down-sampling. Last but not least, a fixed time window like the one proposed in~\cite{he2014practical} would allow us to obtain most of the positives before training, but would still lead to FN labels for those engagements that fall outside the fixed window. Consequently, the problem of handling false negatives would still persist with the additional risk of the model becoming stale.

In this work we devise a model that predicts the probability (pCTR) that a video ad will be clicked by a particular user on the Twitter platform. Through this endeavour, we compare two different classes of model architectures and five different loss functions offline on two different datasets. Subsequently, we pick the top performing architecture and loss functions and evaluate them through online experiments. The first model architecture is a simple logistic regression model that has been extensively used in the industry of display advertising~\cite{he2014practical,chapelle2014modeling} due to its simplicity, good performance and ease of ingesting and handling new training examples with online training. The second model employs a wide \& deep architecture~\cite{cheng2016wide} and was introduced to tackle the complexity and diversity of features used in recommender systems. The five loss functions we test are log loss, FN weighted loss, FN calibration, positive unlabeled loss \cite{du2014analysis,du2015convex} as well as delayed feedback loss \cite{chapelle2014modeling}. Among these, the log loss has been commonly used for CTR prediction~\cite{juan2016field}. Continuous learning is used during the online experiments, which we consider to offer the best trade-off in terms of infrastructure cost, ease of productionization and has the significant advantage that the model is constantly trained on fresh data.

While continuous or online training via online gradient descent is widely used~\cite{hazan2008adaptive} in shallow (linear or kernel-based) models, there is relatively little work on continuous learning of deep neural networks~\cite{liu2017pbodl}. This is particularly challenging, due to the fact that the objective function is non-convex and alternative approaches, like a hedging strategy~\cite{sahoo2018online}, have been proposed to tackle this problem. To the best of our knowledge, this is the first time a deep learning model is used to estimate pCTR in display advertising while addressing the issue of delayed feedback. Given that online training is challenging to adopt with deep neural networks, this work aims to benchmark and suggest feasible solutions to the issue of delayed feedback without additional engineering cost. Three of the loss functions under consideration, positive unlabeled (PU), FN weighted and FN calibration, are applied for the first time on this problem, while the last two are based on importance sampling. Our results indicate that good performance with linear models and small datasets does not necessary translate to equivalent performance with deep models. Indicatively, the delayed feedback loss leads to the best RCE using a linear model on the public dataset, but is outperformed by all proposed loss functions with a deep model on Twitter data offline (2.99\% RCE increase). The efficacy of the loss functions also changes with the amount of data available. 
We hope that this paper can serve as a guideline for which loss function to use when training deep learning models in a continuous fashion for the task of ad-click prediction.

\section{Related Work}
\label{sec:related_work}
In order to ensure that a model is constantly trained on a fresh stream of data, some examples are falsely labeled as negatives or remain unlabeled. As soon as the user engages with the ad, the same data point will be presented to the model with a positive label. In~\cite{he2014practical} they claim to use a time window that is sufficiently long to significantly reduce the bias between the empirical CTR and the ground truth (arising from a fraction of impressions falsely being labeled as negative). Even though most approaches disregard the time delay information that is available (i.e. time that has elapsed since the impression and time until the user engages with the ad), some of them leverage time-delay by jointly training a delay model along with a CPC or CPA model~\cite{chapelle2014modeling,yoshikawa2018nonparametric}. In the following sections, we describe five different methods, which constitute potential solutions to the FN problem. We further discuss their individual challenges. 

\subsection{Importance Sampling}
The cross-entropy of a model with respect to a data distribution $p$ is given by:

\begin{equation}
    \mathcal{L}(\theta) = -\mathbb{E}_p[\log f_{\theta}(\mathbf{y}|\mathbf{x})] = -\sum_{\mathbf{x}, \mathbf{y}}p(\mathbf{x},\mathbf{y})\log f_{\theta}(\mathbf{y}|\mathbf{x})
    \label{eq:cross_entropy}
\end{equation}

where $\mathbf{x}$ are the features related to a particular request (user and ad related), $\mathbf{y}$ is a binary label representing engagement and $f_\theta$ is the model that tries to estimate $p$. As previously discussed, in online training scenarios, samples are introduced as negative examples until a positive label is observed. That leads the model to observe a biased distribution, $b$, instead of the actual data distribution. Hence, we cannot sample from $p$ but only have access to samples from a different distribution $b$. With the application of an appropriate weighting scheme we can obtain an unbiased estimate of the expectation in eq.~\ref{eq:cross_entropy} by using:

\begin{equation}
    \mathbb{E}_p[\log f_{\theta}(\mathbf{y}|\mathbf{x})]=\mathbb{E}_b\big[\frac{p(\mathbf{x},\mathbf{y})}{b(\mathbf{x},\mathbf{y})}\log f_{\theta}(\mathbf{y}|\mathbf{x})\big]
    \label{eq:expectation}
\end{equation}

The weights $w(\mathbf{x},\mathbf{y})=\frac{p(\mathbf{x},\mathbf{y})}{b(\mathbf{x},\mathbf{y})}$ correspond to the importance weights and aim to correct for the fact that the averaging is performed over a different distribution. Using samples from a different, biased distribution $b$, the expectation in eq.~\ref{eq:expectation} can be estimated with:

\begin{equation*}
    \frac{1}{N}\sum_n w(\mathbf{x}_n,\mathbf{y}_n) \log f_{\theta}(\mathbf{y}_n|\mathbf{x}_n)
\end{equation*}

This method of using samples from one distribution to estimate the expectation with respect to another distribution is called \textit{importance sampling}. Importance sampling has been extensively used in different contexts, including counterfactual analysis, and in~\cite{bottou2013counterfactual} authors discuss the assumptions involved and how this technique can be applied in computational advertising to estimate the counterfactual expectation of any quantity. The biggest challenge of using importance sampling in this setting is that we need an estimate of the weights $w(\mathbf{x},\mathbf{y})$. From this point onward, we use $f_{\theta}(\mathbf{x})$ to refer to $f_{\theta}(\mathbf{y}=1|\mathbf{x})$ for brevity.

\subsection{Inverse Propensity Weighting}
Inverse propensity weighting has been widely used in causal inference, where certain samples within the population might be underrepresented. The discrepancy between the actual population and the sample can be adjusted for by using appropriate weights on the individual samples in the estimator~\cite{austin2015moving}. When the sampling probability is known, then the inverse of this probability is used to weight the observations. The \textit{propensity score}, denoted as $p(x)=P(T=1|X=x)$, is the probability that a sample will be assigned a particular treatment given a set of covariates. Under the assumption that the treatment is not randomly assigned, the counterfactual is the equivalent estimation as if all samples in the population could be assigned either treatment with equal probability. A similar approach can be used in the context of click-through rate prediction where treatment is the equivalent of an ad being assigned a ground truth label. One requirement that needs to be taken into account when applying this technique in any problem is that the propensity weights need to be estimated by a separate, reference model.

\subsection{Positive - Unlabeled Learning}
Another set of approaches does not explicitly identify data with negative labels and only learns from positive (P) and unlabeled (U) examples, unlike the traditional classification setting where both positive and negative examples are available. This setting arises when it is impossible or very costly to obtain labeled examples, e.g. in text classification~\cite{fung2006text}, molecular biology~\cite{elkan2008learning} but has also been applied for outlier detection~\cite{scott2009novelty} and time series classification~\cite{nguyen2011positive}. In this scenario, the available training data consists of an incomplete, but randomly sampled, set of positive examples and a set of unlabeled examples, which can either be positive or negative, leading to a need for a different training procedure. A key assumption about the training data is that they are drawn randomly from $p(\mathbf{x}, \mathbf{y}, \mathbf{s})$, and for each tuple $<\mathbf{x}, \mathbf{y}, \mathbf{s}>$ that is drawn, only $<\mathbf{x}, \mathbf{s}>$ is recorded. Here $\mathbf{s}$ is the observed label and $\mathbf{y}$ is the actual label, which might not have occurred yet. Along with this, it is assumed that labeled positive examples are chosen completely randomly from all positive examples, i.e. $p(\mathbf{s}=1|\mathbf{x}, \mathbf{y}=1) = p(\mathbf{s}=1|\mathbf{y}=1)$.

In this case, a classifier $g(\mathbf{x})=p(\mathbf{s}=1|\mathbf{x})$ can be first trained to estimate the probability that an example is labeled positive ($\mathbf{s}=1$) given that its original label is positive ($\mathbf{y}=1$)~\cite{elkan2008learning}, since 
\begin{equation*}
    p(\mathbf{s}=1|\mathbf{y}=1)=\frac{1}{n_P} \sum_{\mathbf{x} \in P} g(\mathbf{x})
\end{equation*}

where $n_P$ is the cardinality of positives P. In the next step, each unlabeled example can be treated as a combination of a positive example with weight proportional to $p(\mathbf{y}=1|\mathbf{x}, \mathbf{s}=0)$ and a negative example with a complementary weight $1-p(\mathbf{y}=1|\mathbf{x}, \mathbf{s}=0)$, while all positive examples have a unit weight. This weight can be expressed as:

\begin{equation*}
    p(\mathbf{y}=1|\mathbf{x}, \mathbf{s}=0)=\frac{1-p(\mathbf{s}=1|\mathbf{y}=1)}{p(\mathbf{s}=1|\mathbf{y}=1)}\frac{p(\mathbf{s}=1|\mathbf{x})}{1-p(\mathbf{s}=1|\mathbf{x})}
\end{equation*}

Finally, a classifier can be trained on the available data using these weights and the standard training procedure. 

Previously,~\cite{lee2003learning} learned the conditional probability of observing a positive label given the input by performing logistic regression and optimizing for the sum of squared weights of the linear function and the sum of weighted logit losses. In this setting, unlabeled samples have a unit weight, while positive samples are weighted by the ratio $n_U / n_P$. The PU problem has also been addressed by aggregating classifiers that are trained to discriminate between P data and a random subsample of U data~\cite{mordelet2014bagging}.

In~\cite{du2014analysis} and~\cite{du2015convex}, du Plessis et al. were the first to propose an unbiased non-convex risk estimator and analyzed the excess risk when the class prior is estimated from the data. The classification risk in a standard binary classification setting is given by:

\begin{multline*}
    \hat{R}_{PN}(f_{\theta})= p(\mathbf{y}=1)\mathbb{E}_{p(\mathbf{x}|\mathbf{y}=1)}\big[l(f_{\theta}(\mathbf{x}))\big] + \\ 
    p(\mathbf{y}=0)\mathbb{E}_{p(\mathbf{x}|\mathbf{y}=0)}\big[l(1-f_{\theta}(\mathbf{x}))\big]
\end{multline*}

where $l$ is the loss function.
Since $p(\mathbf{x}) = p(\mathbf{y}=1)p(\mathbf{x}|\mathbf{y}=1) + p(\mathbf{y}=0)p(\mathbf{x}|\mathbf{y}=0)$, the last term can be expressed as:

\begin{multline*}
    p(\mathbf{y}=0)\mathbb{E}_{p(\mathbf{x}|\mathbf{y}=0)}\big[l(1-f_{\theta}(\mathbf{x}))\big] = \\ 
    \mathbb{E}_{p(\mathbf{x})}\big[l(1-f_{\theta}(\mathbf{x}))\big] - p(\mathbf{y}=1)\mathbb{E}_{p(\mathbf{x}|\mathbf{y}=1)}\big[l(1-f_{\theta}(\mathbf{x}))\big]
\end{multline*}

Therefore, the classification risk can be approximated with the following expression:

\begin{multline}
    \hat{R}_{PU}(f_{\theta}) = p(\mathbf{y}=1)\mathbb{E}_{p(\mathbf{x}|\mathbf{y}=1)}\big[l(f_{\theta}(\mathbf{x}))\big] -  \\
    p(\mathbf{y}=1)\mathbb{E}_{p(\mathbf{x}|\mathbf{y}=1)}\big[l(1-f_{\theta}(\mathbf{x}))\big] + \mathbb{E}_{p(\mathbf{x})}\big[l(1-f_{\theta}(\mathbf{x}))\big]
    \label{eq:risk_pu}
\end{multline}

where $p(\mathbf{x})$ corresponds to the marginal distribution over all unlabeled examples. According to~\cite{kiryo2017positive}, this unbiased risk estimator can yield negative empirical risks if the model trained is too flexible, which makes this loss function more difficult to optimize with neural networks and prone to overfitting. 

\subsection{Delayed Feedback Models}
The approach presented by~\cite{chapelle2014modeling} models a CPA and takes into account the time that has elapsed since the ad click and does not involve a matching window. In this context the issue of delayed feedback is rather addressed as follows: training samples are only labeled as positive if a positive label has actually been observed and are otherwise considered unlabeled, similarly to the PU approach. In the case of modeling post-click attribution, it means that negative labels cannot occur because a conversion can happen at any time in the future. Most approaches that learn only from positive and unlabeled examples~\cite{lee2003learning,elkan2008learning} assume that the probability of having a positive example with a missing label is constant. According to~\cite{chapelle2014modeling}, however, recent clicks are less likely to be assigned true labels because not enough time has elapsed, i.e. the probability of positive examples is time-dependent. Therefore, they use a second model that strongly resembles survival time analysis models~\cite{kalbfleisch2011statistical} to capture the expected delay between click and conversion. This is separate to the model that predicts whether the user will eventually convert, but the two models are trained jointly. In this method, the random variable $Y \in \{0, 1\}$ indicates whether a conversion has already occurred, while a separate random variable $C \in \{0, 1\}$ indicates whether the user will eventually convert. Once the two models are trained, only the one that predicts the probability of conversion, i.e. $P(\mathbf{c}=1|\mathbf{x})$, is preserved, while the model of conversion delay $P(d|\mathbf{x},\mathbf{c}=1)$ is discarded. A standard logistic regression model represents the probability of click, while an exponential (non-negative) distribution is assumed for delay, i.e.: 

\begin{equation}
    P(d|\mathbf{x},\mathbf{c}=1)=\lambda(\mathbf{x})\exp (-\lambda(\mathbf{x}) d)
\end{equation}

\begin{figure}[t]
    \centering
    \includegraphics[width=\linewidth]{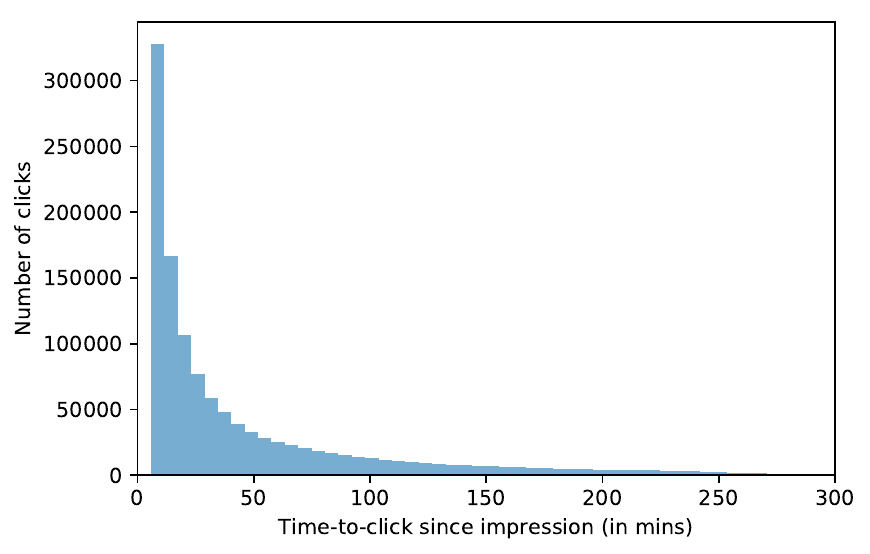}
    \caption{Distribution of time-to-click delay for training ads (longer than 5mins). Corresponds to the distribution after correcting for the CDF of the censoring distribution.}
    \label{fig:exp_dist}
\end{figure}

Hence, $\mathbf{y}=0$ can occur under two different circumstances: either the time elapsed $e$ is shorter than the time to conversion $d$, i.e. $e < d$, or the user will never convert, i.e. $\mathbf{c}=0$. Even though this model was applied in a cost-per-conversion model, it is also applicable in the cost-per-click model that is the main focus of this paper. Fig.~\ref{fig:exp_dist} illustrates that the time-to-click also follows an exponential distribution, which renders this model an appropriate solution. As an extension of the model presented in~\cite{chapelle2014modeling},~\cite{yoshikawa2018nonparametric} suggested a non-parametric delayed feedback model (\textit{NoDeF}) to capture the time delay without assuming any parametric distributions, like exponential or Weibull. This model assumes a hidden variable for each sample, which indicates whether this action will eventually lead to conversion. For the parameter estimation, the expectation-maximization (EM) algorithm~\cite{dempster1977maximum} is used. Adopting these approaches would, however, require estimating the time delay with a separate model and significantly increase the infrastructure cost and complexity to put such a system into production.

\subsection{Delayed Bandits}
Delayed feedback has been extensively considered in the context of Markov Decision Processes (MDPs)~\cite{katsikopoulos2003markov,walsh2009learning}, but their application becomes more challenging for unbounded delays. In~\cite{vernade2017stochastic} authors propose discrete-time stochastic multiarmed bandit models to address the problem of long delays with possibly censored feedback, i.e. feedback that is not observable. They cover two different versions of these models: the \textit{uncensored} model which allows conversions to be observed after an arbitrarily long delay, and the \textit{censored} model which imposes a window restriction of $m$ time steps following the action, after which a conversion cannot be observed anymore. At each round the agent receives a reward that corresponds to the number of observed conversions at time $t$. However, unlike the previous approach, these models operate under the assumption that the distribution delay is known. They consider this to be a valid assumption since the distribution can be estimated from historical data and also claim that delay distribution can be estimated in an online fashion at no additional cost, under the assumption that it is shared by all actions. 

\section{Proposed Approach}

In this work we adopt a continuous training scheme, which suggests that we can potentially wait infinite time since the ad impression until a positive engagement is eventually observed. Historically, this has been dealt with by ingesting all samples with a negative label until a positive engagement is recorded by the user. Hence, the biased data distribution, which is the observed distribution, contains all samples from the actual data distribution labeled as negatives. Accordingly, the positive examples in the biased distribution correspond to all positives from the original data distribution.

\subsection{Model Architecture}
This section describes the architectural details for the models under consideration.

\subsubsection{Logistic regression}
We use a standard logistic regression model, which has been used very extensively in the field of display advertising~\cite{he2014practical,chapelle2014modeling}:

\begin{equation*}
    f_{\theta}(\mathbf{x}) = \frac{1}{1+\exp(-\mathbf{w}_c \cdot \mathbf{x})} = \sigma(\mathbf{w}_c \cdot \mathbf{x})
\end{equation*}

$\sigma(\cdot)$ corresponds to the sigmoid function, while the input $\mathbf{x}$ is a sparse representation of thousands of features related to the users and the ad candidates for a particular request.

\subsubsection{Wide-and-deep model}
This deep model consists of a wide component, corresponding to a generalized linear model, and a deep component, corresponding to a standard feed-forward neural network. The wide component handles both raw input features and transformed, e.g. cross-product features, adding non-linearity to the generalized linear model. The deep component converts high-dimensional sparse features to an embedding vector, transforming categorical features to a dense, low-dimensional representation. 

Similarly to the previous model, features include user features, contextual features and ads features. The model's prediction for CTR is given by:

\begin{equation*}
    f_{\theta}(\mathbf{x})= \sigma(\mathbf{w}_{wide}^T[\mathbf{x}, \phi(\mathbf{x})] + \mathbf{w}_{deep}^T \alpha^{(l_f)} + b)
\end{equation*}

$\mathbf{w}_{wide}$ corresponds to the weights of the wide component, $\mathbf{w}_{deep}$ the weights of the deep component, $\phi(\mathbf{x})$ the cross-product transformations and $\alpha^{(l_f)}$ the final layer activations of the deep branch.

\subsection{Loss Functions}
\subsubsection{Delayed feedback loss}
In this version of the loss function, an exponential distribution is assumed for the time delay and this model is jointly trained with either the logistic regression or the deep model. Assuming that $\lambda(\mathbf{x})=\exp(\mathbf{w}_d \cdot \mathbf{x})$ and $\mathbf{w}_c$ the parameters of the pCTR model, we optimize for the regularized log likelihood with respect to the parameters $\theta$ and $\mathbf{w}_d$:

\begin{equation*}
    arg \min_{\theta, \mathbf{w}_d} \mathcal{L}_{DF}(\theta, \mathbf{w}_d) + \alpha (\| \theta \|^2_2 + \|\mathbf{w}_d\|^2_2)
\end{equation*}

where $\alpha$ is the regularization parameter and $\mathcal{L}_{DF}$ is:

\begin{multline}
    \mathcal{L}_{DF}(\theta, \mathbf{w}_d) = - \sum_{\mathbf{x}, \mathbf{y=1}} \log f_{\theta}(\mathbf{x}) + \log \lambda(\mathbf{x}) - \lambda(\mathbf{x})d \\
    - \sum_{\mathbf{x}, \mathbf{y=0}} \log[1 - f_{\theta}(\mathbf{x}) + f_{\theta}(\mathbf{x}) \exp(-\lambda(\mathbf{x})e)]
\end{multline}

$f_{\theta}(\mathbf{x})$ corresponds to the output of the pCTR model, $d$ corresponds to the time-to-click for a positive example, while $e$ represents the time elapsed since the ad impression.
This loss function can be computed in a more numerically stable way as follows:

\begin{multline*}
    \mathcal{L}_{DF}(\theta, \mathbf{w}_d) = - \sum_{\mathbf{x}, \mathbf{y}} \log f_{\theta}(\mathbf{x}) - \sum_{\mathbf{x}, \mathbf{y=1}} \mathbf{w}_d \cdot \mathbf{x} - \lambda(\mathbf{x})d \\
    - \sum_{\mathbf{x}, \mathbf{y=0}} \log[\exp(-f_{\theta}(\mathbf{x})) +  \exp(-\lambda(\mathbf{x})e)]
\end{multline*}

\subsubsection{Positive-unlabeled loss}
In this section, we consider using the PU loss under the FN setting by treating all negative samples in the biased training data as unlabeled. According to eq.~\ref{eq:risk_pu}, the following loss function can be derived:

\begin{equation}
    \mathcal{L}_{PU}(\theta) = -\sum_{\mathbf{x}, \mathbf{y}=1} [\log f_{\theta}(\mathbf{x}) - \log (1 - f_{\theta}(\mathbf{x}))] - \sum_{\mathbf{x}, \mathbf{y}=0} \log (1 - f_{\theta}(\mathbf{x}))
\end{equation}

Empirically this could be perceived as applying the traditional log loss for both positive and negative samples. In addition, a step in the opposite direction of negative gradients is made whenever a positive example is observed. This assumption is reasonable given the fact that for every positive sample there have been parameter updates based on gradients from a fake negative sample.

\subsubsection{Fake negative weighted}
This loss relies on importance sampling. In our training setting, samples are labeled as negatives and ingested to the training pipeline, and then duplicated with a positive label as soon as a user engagement takes place. To formulate this loss function, we rely on the following assumptions: $b(\mathbf{x}|\mathbf{y}=0) = p(\mathbf{x})$ and $b(\mathbf{x}|\mathbf{y}=1)=p(\mathbf{x}|\mathbf{y}=1)$, where $b$ is the biased observed distribution and $p$ is the actual data distribution. We also know that $b(\mathbf{y}=0)=\frac{1}{1+p(\mathbf{y}=1)}$, since all samples are initially labeled as negative.

The loss function in (\ref{eq:cross_entropy}) can be written as:

\begin{multline}
        -\sum_{\mathbf{x}, \mathbf{y}}p(\mathbf{y}=1|\mathbf{x})\log f_{\theta}(\mathbf{x}) + p(\mathbf{y}=0|\mathbf{x})\log f_{\theta}(\mathbf{y}=0|\mathbf{x})  = \\
        -\sum_{\mathbf{x}, \mathbf{y}}  b(\mathbf{y}=1|\mathbf{x})\frac{p(\mathbf{y}=1|\mathbf{x})}{b(\mathbf{y}=1|\mathbf{x})}\log f_{\theta}(\mathbf{x}) + \\
        b(\mathbf{y}=0|\mathbf{x})\frac{p(\mathbf{y}=0|\mathbf{x})}{b(\mathbf{y}=0|\mathbf{x})}\log f_{\theta}(\mathbf{y}=0|\mathbf{x})
        \label{ml:is_loss}
\end{multline}

The probability of observing a positive user engagement in the biased distribution is:

\begin{equation*}
        b(\mathbf{y}=1|\mathbf{x}) = \frac{b(\mathbf{y}=1)b(\mathbf{x}|\mathbf{y}=1)}{b(\mathbf{y}=1)b(\mathbf{x}|\mathbf{y}=1) + b(\mathbf{y}=0)b(\mathbf{x}|\mathbf{y}=0)}
\end{equation*}

which, using the above assumptions and assigning $w(\mathbf{x}) \coloneqq \frac{1}{1 + p(\mathbf{y}=1|\mathbf{x})}$, can be expressed as:
\begin{multline}
     b(\mathbf{y}=1|\mathbf{x}) =\frac{w(\mathbf{x})p(\mathbf{y}=1)p(\mathbf{x}|\mathbf{y}=1)}{w(\mathbf{x})p(\mathbf{y}=1)p(\mathbf{x}|\mathbf{y}=1) + w(\mathbf{x})p(\mathbf{x})} = \\ 
     \frac{p(\mathbf{y}=1|\mathbf{x})p(\mathbf{x})}{p(\mathbf{y}=1|\mathbf{x})p(\mathbf{x}) + p(\mathbf{x})} = \frac{p(\mathbf{y}=1|\mathbf{x})}{1 + p(\mathbf{y}=1|\mathbf{x})}
     \label{eq:b_y_1}
\end{multline}

Similarly, the probability that the user will not engage is the following:

\begin{equation}
    b(\mathbf{y}=0|\mathbf{x})=1-b(\mathbf{y}=1|\mathbf{x})=\frac{1}{1 + p(\mathbf{y}=1|\mathbf{x})}
    \label{eq:b_y_0}
\end{equation}

By replacing (\ref{eq:b_y_1}) and (\ref{eq:b_y_0}) in eq.~\ref{ml:is_loss}, we obtain the following expression:

\begin{multline}
        \mathcal{L}_{IS}(\theta) = -\sum_{\mathbf{x}, \mathbf{y}} b(\mathbf{y}=1|\mathbf{x})(1 + p(\mathbf{y}=1|\mathbf{x}))\log f_{\theta}(\mathbf{x}) + \\
        b(\mathbf{y}=0|\mathbf{x})p(\mathbf{y}=0|\mathbf{x})(1 + p(\mathbf{y}=1|\mathbf{x}))\log f_{\theta}(\mathbf{y}=0|\mathbf{x})
        \label{eq:is_loss}
\end{multline}

Therefore, we can weigh positive samples with $(1 + p(\mathbf{y}=1|\mathbf{x}))$ and negative samples with $(1 - p(\mathbf{y}=1|\mathbf{x}))\cdot(1 + p(\mathbf{y}=1|\mathbf{x}))$. Since we don't directly have access to $p$, we can replace it with the model estimate $f_{\theta}$, as long as $f_{\theta}$ converges to $p$, which we prove in the following paragraphs.

Relying on (\ref{eq:b_y_1}) and (\ref{eq:b_y_0}), and by replacing $p$ with $f_{\theta}$ in the importance weights, (\ref{eq:is_loss}) can be rewritten as:

\begin{multline*}
    \mathcal{L}_{IS}(\theta) = -\sum_{\mathbf{x}, \mathbf{y}}\frac{p(\mathbf{y}=1|\mathbf{x})}{1 + p(\mathbf{y}=1|\mathbf{x})}\big[(1+f_{\theta}(\mathbf{x}))\big]\log f_{\theta}(\mathbf{x}) + \\
    \frac{1}{1 + p(\mathbf{y}=1|\mathbf{x})} \big[(1 - f_{\theta}(\mathbf{x}))(1 + f_{\theta}(\mathbf{x}))\big] \log((1 - f_{\theta}(\mathbf{x}))
\end{multline*}

The terms in $\big[ \cdot \big]$ brackets are not taken into account in the gradient calculation of the loss with respect to the input. Finally, the gradient of $\mathcal{L}_{IS}(\theta)$ with respect to $f_{\theta}$ can be written as:

\begin{multline}
    \frac{\partial \mathcal{L}_{IS}}{\partial f_{\theta}} = - \frac{p(\mathbf{y}=1|\mathbf{x})}{1 + p(\mathbf{y}=1|\mathbf{x})} \frac{1+f_{\theta}(\mathbf{x})}{f_{\theta}(\mathbf{x})} + \frac{1+f_{\theta}(\mathbf{x})}{1 + p(\mathbf{y}=1|\mathbf{x})} \\
    = \frac{(1+f_{\theta}(\mathbf{x}))(f_{\theta}(\mathbf{x})-p(\mathbf{y}=1|\mathbf{x}))}{(1 + p(\mathbf{y}=1|\mathbf{x}))f_{\theta}(\mathbf{x})}
\end{multline}

Note that when $\partial \mathcal{L}_{IS} / \partial f_{\theta}=0$, then $f_{\theta}(\mathbf{x}) = p(\mathbf{y}=1|\mathbf{x})$, i.e. $f_{\theta}(\mathbf{x})$ converges to $p(\mathbf{y}=1|\mathbf{x})$, as long as $f_{\theta}(\mathbf{x}) > 0$. When $f_{\theta}(\mathbf{x}) > p(\mathbf{y}=1|\mathbf{x})$, $\partial \mathcal{L}_{IS} / \partial f_{\theta}>0$, and when $f_{\theta}(\mathbf{x}) < p(\mathbf{y}=1|\mathbf{x})$, $\partial \mathcal{L}_{IS} / \partial f_{\theta}<0$, for $p(\mathbf{y}=1|\mathbf{x}) \in (0, 1]$. This indicates that the FN weighted loss leads to $f_{\theta}(\mathbf{x})=p(\mathbf{y}=1|\mathbf{x})$ and the gradients always point towards the right direction.

\subsubsection{Fake negative calibration}
In this approach the model estimates the biased distribution $b$ and, then, the following transformation is used after solving eq.~\ref{eq:b_y_1} for $p(\mathbf{y}=1|\mathbf{x})$:

\begin{equation*}
    p(\mathbf{y}=1|\mathbf{x}) = \frac{b(\mathbf{y}=1|\mathbf{x})}{1-b(\mathbf{y}=1|\mathbf{x})}
\end{equation*}

This is always a valid distribution, since for every positive in the biased distribution a FN is observed, i.e. $b(\mathbf{y}=1|\mathbf{x}) \leq 0.5$ and $p(\mathbf{y}=1|\mathbf{x}) \leq 1$. For the sake of brevity, we refer to this method as FN calibration.

\section{Experiments}
\subsection{Setup}
\subsubsection{Offline metrics.}
In order to compare the different model architectures and loss functions under consideration, we first performed offline experiments. The offline experiments were used to validate that the proposed loss functions are appropriate for the problem of CTR prediction and two different datasets, one in-house and one public dataset, were used for training and testing the models. In the offline setting the metrics that we focus on and report in this paper include \textit{log loss} on the evaluation set (which does not contain fake negatives), \textit{relative cross entropy} (RCE) and \textit{area under precision-recall curve} (PR-AUC). RCE corresponds to the improvement of a prediction relative to the straw man, or the naive prediction, measured in cross entropy (CE). The naive prediction corresponds to the case that does not take into account the user and ad features, e.g. it always predicts the average CTR of the training set. Suppose the average CE of the naive prediction is $CE_{naive}$ and average CE of the prediction to be evaluated is $CE_{pred}$, then RCE is defined as $(CE_{naive}-CE_{pred}) * 100 / CE_{naive}$. Note that the lower the CE the better the quality of the predictions, so the higher the RCE. The benefit of using RCE is that we can obtain a confident estimate of whether the model is under or over performing the naive prediction. PR-AUC is a more commonly used metric and is more sensitive than AUC on skewed data.

\subsubsection{Online metrics.}
The model and loss functions with the most promising offline results were further evaluated in an online setting, since the desired application is continuous training. The online setting reflects the actual performance that determines which approach is the most suitable for the problem of delayed feedback. The evaluation is, then, performed on a holdback evaluation dataset of Twitter data to compare performance between the `control` methodology and the `treatment` (in an A/B testing framework). The fake negatives have been removed from this holdback dataset by waiting up to 9 hours following the end of the evaluation date for an engagement label. Two key metrics are reported in this setting: \textit{pooled relative cross entropy} (pooled RCE) and \textit{revenue per thousand requests} (RPMq)~\cite{li2015click}. Pooled RCE is used to address a fair comparison of RCE between control and experiment, since otherwise each of the two models is verified on its own traffic, and is an indicator of how well a model will generalize by evaluating it on the pooled traffic generated by both models. RPMq represents revenue made per 1000 ad requests. Increasing user engagement by showing higher quality ads can cause RPMq to increase, but RPMq can also go up by just serving more ads per request, irrespective of their quality. Therefore, higher RPMq is desirable but this metric needs to be considered along with CTR, since a high RPMq with low CTR could potentially hurt user experience and cause negative long-term effects.

\subsubsection{Hyperparameters.}
The hyperparameters used for the offline experiments are: stochastic gradient descent (SGD) optimizer, learning rate 0.02; decay rate 0.000001; batch size 128; the learning rate used for the delayed feedback loss is 0.005 and the $l_2$ regularisation parameter for the delayed feedback model is 2. We use a mix of categorical and continuous features which are discretized into a fixed number of bins. The deep part of the wide-and-deep model consists of 4 layers with sizes $[400, 300, 200, 100]$ and a leaky rectified linear unit (ReLU) is adopted as an activation function for the intermediate layers. The weights are initialized using Glorot~\cite{glorot2010understanding} initialization. The same hyperparameters were used to initialize the online models.


\subsection{Data}
\begin{table}[t]
\begin{tabular}{lrrr}
\hline
\multicolumn{4}{c}{Logistic regression - Criteo data}        \\ \cline{1-4}
Loss function   & Loss   & RCE & PR-AUC \\ \hline
Log loss   &  0.3963   &  17.26   &  \textbf{0.5081}  \\
Delayed feedback loss  &  0.3970  & \textbf{17.32}  &  0.5080  \\ 
PU loss  &  0.4065   &  15.10   &  0.5048   \\
FN weighted   &  0.4008   &  16.30  &  0.5037  \\
FN calibration   &   \textbf{0.3961}   &  17.29    &   0.4983   \\
\hline
\end{tabular}
\caption{Offline results with the linear model on public Criteo data. Bold indicates top-performing method. Differences between top and second-best method are not statistically significant.}
\label{tab:offline_criteo}
\end{table}

\subsubsection{Public data.}
The public dataset used to evaluate the different loss functions offline was provided by Criteo~\cite{chapelle2014modeling}. It should be mentioned that this dataset corresponds to conversions after a click has occurred, so the time delay is generally longer in comparison to the CTR prediction. Each sample is described by a set of hashed categorical features and a few continuous features. The total number of training data amounts to 15.5 million examples, while the evaluation set consists of 3.5 million samples. In order to create a dataset with fake negatives from the original public dataset the following process is carried out: the latest conversion time or click time is used as snapshot time and a fake negative example is introduced at the time of click for all positive examples. The original dataset is used to evaluate logistic and delayed feedback loss functions with a logistic regression model. The version of the dataset that includes the FN examples at the time of click is used to evaluate the PU, FN weighted and FN calibration loss functions.

\subsubsection{Offline Twitter data.} For the experiments based on in-house data, we train on 4 days of data offline. The evaluation is performed on the following day's data.
Given that only a small fraction of the ads served to the users are actually clicked, data label imbalance poses a particular challenge. In our training setup the negative examples are downsampled to 5\% of the original dataset to address this imbalance problem and a higher weight is adopted for negative samples in the loss function to account for this modification. After these steps the total number of training data amounts to 668 million video ads, while the test data amounts to 7 million ads.
For the evaluation dataset, a positive label is assigned to a sample if an engagement takes place up to 9 hours following the impression time. Otherwise, samples are assigned a negative label. Both positive and negative examples are, subsequently, downsampled to 5\% of their original size. In order to obtain the time elapsed and time-to-click information, a snapshot of the data is captured at the end of each date. Hence, for samples that an engagement has not been observed yet a negative label is assigned and the time elapsed from the impression time until the snapshot time is added as a feature. For samples with observed engagement prior to the snapshot time, apart from the time elapsed, the difference between engagement time and impression time is recorded as the time-to-click. These time features are only used for the estimation of the time delay model which is required by the delayed feedback loss. It is worth mentioning that, since most fake negatives are scrubbed from this derived dataset, the positive / negative ratio is eventually higher than before.

\begin{table}[t]
\begin{tabular}{lrrr}
\hline
\multicolumn{4}{c}{Logistic regression - Twitter data}        \\ \cline{1-4}
Loss function   & Loss   & RCE &  PR-AUC \\ \hline
Log loss   & 0.6087  & 5.48 &   0.5596  \\
Delayed feedback loss  &  0.5725  & 10.59  &    0.5725  \\
PU loss  & 0.5754  & 10.65  &  0.5690  \\
FN weighted   & 0.5713 & 11.30 &   0.5730  \\
FN calibration    &  \textbf{0.5641}$^*$  & \textbf{12.41}$^*$ &  \textbf{0.5813}$^*$  \\ \hline
\end{tabular}
\caption{Offline results with the logistic regression model on Twitter data. Bold indicates top-performing method and (*) statistical significance using an unpaired t-test in comparison to second-best performing method ($p<0.05$).}
\label{tab:offline_log_ABCD}
\end{table}

\subsubsection{Online Twitter data.}
In online experiments, all the models train on a continuous data stream that is generated from impression callback data in real time. Ad impressions are served to users and the label for each example is decided based on current click information (this is where the fake negatives enter the training data). Each training example is then published to a data stream which the model's training service subscribes to. The continuous training process outputs models every 10 minutes which are picked up by the prediction service to serve online traffic.
When computing pooled RCE, we use the same data source that is used to generate offline evaluation data, meaning that we remove fake negatives by waiting for 9 hours before assigning labels to each ad impression. 

It is important to note that each online experiment serves only 1\% of production traffic, meaning that the training traffic is dominated by our current production model. We also compute our online pooled RCE metric on this reference production traffic which is not affected by any of the models, to ensure fairness by using the same evaluation dataset for each model.

\subsection{Results}

\begin{table}[t]
\begin{tabular}{lrrr}
\hline
\multicolumn{4}{c}{Wide \& deep - Twitter data}        \\ \cline{1-4}
Loss function   & Loss   & RCE & PR-AUC \\ \hline
Log loss   &  0.5953  &  7.81  &  0.5872  \\
Delayed feedback loss &  0.5781 & 12.11  &  0.5781  \\ 
PU loss  &  0.5567  &  13.57  &  \textbf{0.5927}  \\
FN weighted   &  0.5568  & 13.54 &   0.5925  \\
FN calibration   &   \textbf{0.5566}  &  \textbf{13.58}  &   0.5923 \\
\hline
\end{tabular}
\caption{Offline results with the wide \& deep model on Twitter data. Bold indicates top-performing method. Differences between top and second-best method are not statistically significant.}
\label{tab:offline_wd_ABCD}
\end{table}

\subsubsection{Offline evaluation.} Offline results on the Criteo dataset are presented in Table~\ref{tab:offline_criteo}. On this dataset the delayed feedback loss yields the highest RCE (17.32), followed by the FN calibration loss (with RCE 17.29), which is almost on par with the log loss. These results align with the ones reported in the original paper~\cite{chapelle2014modeling}, as the evaluation loss when training with the traditional log loss is 0.3963 and for the delayed feedback loss it is 0.3970. The worst performing loss function on the public data is PU loss, which also showed unstable performance across different runs. These results indicate that the delayed feedback loss is more appropriate with a simple pCTR model (like logistic regression) and fewer training examples, while more complex models, like wide \& deep require more robust solutions that leverage importance sampling techniques in a principled way. Additionally, the PU loss empirically seems harder to optimize for a simple model, as the evaluation metrics fluctuate across the different runs.

The results on Twitter data are presented in Tables~\ref{tab:offline_log_ABCD} and~\ref{tab:offline_wd_ABCD} for the logistic regression and the wide \& deep model, respectively. These correspond to median performance across 8 evaluations of each model with different initialization. Overall, we can observe that the deep learning model is performing better than the logistic regression model for all loss functions, as expected. In terms of RCE, the log loss performs the worst for both models while the FN calibration loss leads to the best performance for the linear model (with RCE 12.41 and loss 0.5641). PU loss and FN calibration perform best for the deep model and yield almost equivalent results (RCE 13.57 and 13.58, respectively) and have very similar performance to the FN weighted loss (RCE 13.54). Consequently, these three loss functions are compared to the log loss in an online setting. The delayed feedback loss performs better than the log loss for both classes of models (RCE 10.59 vs 5.48 for the linear model, and 12.11 vs 7.81 for the deep model). PR-AUC may not vary as much as RCE, but its difference between the top-performing methods (using an unpaired t-test between 1st and 2nd best methods) is still statistically significant for the linear model.

\subsubsection{Online evaluation.}
It should be mentioned that the following results correspond to a budget-unaware experiment, i.e. the advertisers' budget is not taken into account when deciding which ads to display for a particular request, that ran for 1 week. Table~\ref{tab:online_ABCD} shows the online results for the top performing loss functions using the wide \& deep model. Both FN weighted and FN calibration yield higher RPMq compared to the traditional log loss (increases of +55.10\% and +54.37\%). Equivalently, for the monetized CTR the increase is also significant (+23.01\% for FN weighted and 23.19\% for FN calibration). We observe that, despite its good offline performance, the PU loss diverges after 2 days and report online metrics prior to its divergence.

\section{Conclusions}
In this paper we have explored the issue of delayed feedback during continuous learning of neural network models with principled loss functions. Even though click distribution shift can be alternatively handled at the feature level, here we aim to address delayed feedback in particular. The extensive comparison in an online and offline setting using both in-house and public data aims to serve as a guideline for properly addressing this issue in display advertising. Two loss functions proposed in this work, FN weighted and FN calibration, when employed along with a wide \& deep model lead to the best offline performance and translate to online gains. PU loss diverges online and we aim to explore non-negative PU learning~\cite{kiryo2017positive} to resolve this instability in the future.

One element that needs to be taken into account for the proposed loss functions is the temporal dependency of the gradients. This means that parameter updates based on the fake negative samples precede the updates from the positive examples. In the future, we would like to extend this work and combine the principle of importance sampling with modeling time delay by assigning time-dependent weights to the training samples. Particular challenges of continuous learning like catastrophic forgetting~\cite{kirkpatrick2017overcoming} and overfitting are outside the scope of this work, but should be further explored in the context of display advertising. More systematic ways to tackle dataset bias in such recommender systems and adopt exploration / exploitation mechanisms~\cite{chen2019top} also need to be investigated.

\begin{table}[]
\begin{tabular}{lrrr}
\hline
\multicolumn{4}{c}{Wide \& deep - Online experiment}        \\ \cline{1-4}
Loss function   & Pooled RCE   & RPMq & Monetized CTR \\ \hline
Log loss   &  7.68  &  100.00  &  100.00  \\
PU loss  &  \textit{12.27}   &    \textit{137.00}   &  \textit{118.59}  \\
FN weighted   &  \textbf{13.39}  &  \textbf{155.10}  &   123.01  \\ 
FN calibration   &  13.37  &  154.37  &  \textbf{123.19}   \\ \hline
\end{tabular}
\caption{Online results with the wide \& deep model and the best performing loss functions (Twitter data). For RPMq and monetized CTR presented results correspond to relative improvements with respect to the log loss. Results for PU loss are prior to its divergence (within 2 days).}
\label{tab:online_ABCD}
\end{table}

\bibliographystyle{ACM-Reference-Format}
\bibliography{sample-base}

%
\appendix

%
\section*{Supplementary Material}

\quad Stefan Webb (\textit{swebb@twitter.com}) 

Caojin Zhang (\textit{caojinz@twitter.com}) 

Sofia Ira Ktena (\textit{iraktena@gmail.com}) 

Wenzhe Shi (\textit{wshi@twitter.com})

\subsection*{Introduction}
In this supplementary material, we provide a detailed derivation for the fake negative loss. Furthermore, we provide additional experiments that demonstrate the superior performance of fake negative loss compared to the conventional way of addressing the delayed feedback problem.
\section{Derivation of Fake Negative Loss}
This is a short note to derive a Monte Carlo estimate for the fake negative importance sampling loss.
\subsection{Notations}
Let $y$ be the binary-valued random variable representing whether a user performs some ``conversion'' action, such as clicking on a video ad. Let $x$ represent all the relevant features to be included in the prediction model. Then $p(y|x)$ is the probability of conversion, and the objective is to learn a
classifier, $f_{\theta}(y|x)$, via supervised learning.
We want to minimize the cross entropy between $p(y|x)$ and $f_{\theta}(y|x)$ averaged over $p(x)$. Note that this is equivalent to minimizing the KL-divergence between $p(y|x)$ and $f_{\theta}(y|x)$ averaged over $p(x)$. The average cross entropy is given by,
\begin{eqnarray}\label{Loss}
\mathcal{L}(\theta) &=& E_{p(x)} [E_{p(y|x)} [-\ln f_{\theta}(y|x) ]] \\
 &=& - E_{p(x,y)} [\ln f_{\theta}(y|x)]
\end{eqnarray}
that is, Eq (2) in the main paper.
\subsection{Importance Sampling}
The problem, however, is that we do not draw samples from the true data distribution $p$, but rather a biased distribution $b$ that includes ``fake negative'' samples, and so cannot estimate Eq (\ref{Loss}) directly.
We can circumvent this problem by using importance sampling,
\begin{equation}
    \mathcal{L}(\theta) = - E_{b(x,y)} [\frac{p(x,y)}{b(x,y)} \ln f_{\theta}(y|x)]
\end{equation}
This has solved the problem of not being able to draw samples from p(x; y), but now the problem is that we cannot evaluate the density ratio $p(x, y)=b(x, y)$.
Eq (7) in the main paper does not follow from Eq (2) as it has used

\begin{equation*}
    - E_{P(y|x)} [\frac{p(y|x)}{b(y|x)} \ln f_{\theta}(y|x)] \not= \mathcal{\theta}
\end{equation*}

For one thing, the left hand side of the equation does not have any expectation taken with respect to a distribution on $x$ (i.e. it's a random variable).

\subsection{Monte Carlo estimate}
We will make the following assumptions about the biased distribution motivated by the definition of being a fake negative sample,
\begin{eqnarray*}
b(x|y=0)&=&p(x) \\
b(x|y=1)&=&p(x|y=1) \\
b(y=1)&=&\frac{p(y=1)}{1+p(y=1)}
\end{eqnarray*}
Substituting into (\ref{Loss}) allows us to circumvent the second problem mentioned above. First let us expand the expectation,
\begin{eqnarray*}
\mathcal{L}(\theta) &=& - \sum_x \sum_{y\in \{0,1\}} \frac{p(x,y)}{b(x,y)} \ln f_{\theta}(y|x) \\
&=& -\sum_x (\frac{p(x,y=1)}{b(x,y=1)} \ln f_{\theta}(y=1|x) \\ 
&+& \frac{p(x,y=0)}{b(x,y=0)} \ln f_{\theta}(y=0|x))
\end{eqnarray*}
We will simplify the ratio terms using our assumptions. First
\begin{eqnarray*}
b(x,y=1) &=& b(y=1) b(x|y=1) \\
&=&  \frac{p(y=1)}{1+p(y=1)} b(x|y=1) \\
&=& \frac{p(x,y=1)}{1+p(y=1)} \\
&=& b(y=0) p(x, y=1)
\end{eqnarray*}
For the second ratio, we have
\begin{eqnarray*}
b(x,y=0) &=& b(y=0)p(x|y=0) \\
&=& b(y=0) p(x)
\end{eqnarray*}
Substituting them into the loss (\ref{Loss}),
\begin{eqnarray*}
\mathcal{L}(\theta) &=& -\sum_x [\frac{b(x,y=1)}{b(y=0)} \ln f_{\theta}(y=1|x) \\ 
 &&+ p(y=0) \frac{b(x,y=0)}{b(y=0)} \ln f_{\theta}(y=0|x)] \\
&=& -\sum_x [b(x|y=1) \frac{b(y=1)}{b(y=0)} \ln f_{\theta}(y=1|x)  \\ 
&&+ b(x|y=0) p(y=0|x) \ln f_{\theta}(y=0|x) ] \\
&=& \frac{b(y=1)}{b(y=0)} E_{b(x|y=1)} -\ln f_{\theta}(y=1|x)  \\
&&+E_{b(x|y=0)} [-p(y=0|x) \ln f_{\theta}(y=0|x)]
\end{eqnarray*}
Hence a Monte Carlo estimate is,
\begin{eqnarray*}
\hat{\mathcal{L}}(\theta) &=& \frac{N_{y'=1}/N}{N_{y'=0}/N} \frac{1}{N_{y'=1}} \sum_{\{x', y'=1\}} -\ln f_{\theta}(y=1|x') \\
&+& \frac{1}{N_{y'=0}} \sum_{\{x'|y'=0\}} -p(y=0|x') \ln f_{\theta}(y=0|x') \\
&\propto& \sum_{\{x', y'=1\}} -\ln f_{\theta}(y=1|x') \\ 
&&+ \sum_{\{x'|y'=0\}}  -p(y=0|x') \ln f_{\theta}(y=0|x') \\
&\simeq& \sum_{\{x', y'=1\}} -\ln f_{\theta}(y=1|x') \\ 
&&+ \sum_{\{x'|y'=0\}}  -p(y=0|x') \ln f_{\theta}(y=0|x')
\end{eqnarray*}
So to perform approximate importance sampling, we weight positive samples with 1 and negative samples with $1-f_{\theta}(y=1|x')$. This differs from the algorithm in the paper (3.2.3), which weights positive samples with $1+f_{\theta}(y=1|x'))$, and negative samples with $(1 - f_{\theta}(y=1| x=0))(1+f_{\theta}(y =
1|x=0))$.

\section{Supplementary Experiments}
One conventional way to address the delayed feedback problem is to use an attribution window to generate the label for samples. When the impression happens, we can wait $N$ hours to determine whether the sample has a positive engagement label or not. This waiting period is the length of the attribution window. The longer the attribution window, the more accurate the label at the cost of the freshness of the data.

In the original paper, we didn't compare the fake negative loss with log likelihood with different attribution windows. We provide supplemental experiments to demonstrate how fake negative loss can out perform the attribution window approach in this section.

\subsection{Evaluation Metrics} We perform offline experiments to compare the prediction accuracy of different loss function and attribution windows. For prediction accuracy, we choose relative cross entropy (RCE) and precision-recall AUC (PR-AUC).

\subsection{Data \& Models} 
We evaluate our method on ads prediction production service. The baseline model architecture is based on the wide and deep model~\cite{cheng2016wide}.
The training data comprises of 2 days' data, while the following 4 hours are used for testing. To compare the fake negative loss with log likelihood loss, we vary the attribution window for the training data label. The test data uses a 9-hour attribution window to generate the label.

\begin{table}[t]
\begin{tabular}{lrrr}
\hline
\multicolumn{4}{c}{Wide and Deep model - Twitter data}        \\ \cline{1-4}
Loss function   & Attribution   & RCE &  PR-AUC \\ \hline
Log loss   & 1 Hour & $8.02\pm0.03$ &   $0.026 \pm 10^{-4}$  \\
Log loss  &  9 Hours  & $11.21 \pm 0.034$   &    $0.029 \pm 10^{-4}$  \\
Fake negative  & None  & $11.47 \pm 0.029$  &  $0.0303 \pm 10^{-4}$  \\
 \hline
\end{tabular}
\caption{Experiment for various attribution windows and loss functions..}
\label{tab:offline_log}
\end{table}

\section*{Conclusion}
We provide here a detailed derivation for the fake negative loss. Furthermore, our experiments demonstrate that the fake negative loss outperforms the conventional way of addressing the delayed feedback problem. With 1-hour attribution window, the abundance of fake negative samples hurts model performance. When we use longer attribution windows, we obtain more accurate labels and model performance dramatically improves. The fake negative loss not only addresses the delayed feedback problem, but decreases the cost on data freshness that results from longer attribution windows. Hence, model performance can improve further.

\end{document}